%% file: main.tex
\documentclass[10pt,twocolumn,letterpaper]{article}

\usepackage{cvpr}
\usepackage{times}
\usepackage{epsfig}
\usepackage{graphicx}
\usepackage{amsmath}
\usepackage{amssymb}
\usepackage{microtype}
\usepackage{epigraph}
\usepackage{booktabs}
\usepackage{authblk}
\makeatletter
\renewcommand\AB@affilsepx{ \protect\Affilfont}
\makeatother

\newcommand{\TODO}[1]{\textbf{TODO:} \emph{\textcolor{red}{#1}}}

\usepackage{calc}
\newsavebox\CBox
\newcommand\hcancel[2][0.5pt]{%
  \ifmmode\sbox\CBox{$#2$}\else\sbox\CBox{#2}\fi%
  \makebox[0pt][l]{\usebox\CBox}%
  \rule[0.5\ht\CBox-#1/2]{\wd\CBox}{#1}}

\usepackage{enumitem}
\setitemize[0]{leftmargin=10pt}

\newenvironment{tight_itemize}{
\begin{itemize}[leftmargin=8pt]
  \setlength{\topsep}{0pt}
  \setlength{\itemsep}{0pt}
  \setlength{\parskip}{0pt}
  \setlength{\parsep}{0pt}
}{\end{itemize}}

\usepackage[usenames, dvipsnames]{color}

\usepackage[pagebackref=true,breaklinks=true,letterpaper=true,colorlinks,bookmarks=false]{hyperref}

\cvprfinalcopy 

\def\cvprPaperID{122} 

\ifcvprfinal\fi
\begin{document}

\title{\vspace{-4mm}DESIRE: Distant Future Prediction in Dynamic Scenes with Interacting Agents}

\author[1]{Namhoon Lee}
\author[2]{Wongun Choi}
\author[2]{Paul Vernaza}
\author[3]{Christopher B. Choy}
\author[1]{\authorcr Philip H. S. Torr}
\author[2,4]{Manmohan Chandraker}
\affil[1]{University of Oxford, }
\affil[2]{NEC Labs America, }
\affil[3]{Stanford University, }
\affil[4]{University of California, San Diego}

\renewcommand\Authands{, }


\maketitle
\thispagestyle{empty}

\begin{abstract}
    \input{abstract}

\end{abstract}


\input{introduction}

\input{related}

\input{method}

\input{experiments}

\input{conclusion}

{\small
\bibliographystyle{ieee}
\bibliography{egbib}
}

\end{document}

%% file: abstract.tex
We introduce a Deep Stochastic IOC\footnote{IOC: Abbreviation for inverse optimal control, which will be more explained throughout the paper.} RNN Encoder-decoder framework, DESIRE, for the task of future predictions of multiple interacting agents in dynamic scenes. DESIRE effectively predicts future locations of objects in multiple scenes by 1) accounting for the multi-modal nature of the future prediction (i.e., given the same context, future may vary), 2) foreseeing the potential future outcomes and make a strategic prediction based on that, and 3) reasoning not only from the past motion history, but also from the scene context as well as the interactions among the agents. DESIRE achieves these in a single end-to-end trainable neural network model, while being computationally efficient. The model first obtains a diverse set of hypothetical future prediction samples employing a conditional variational auto-encoder, which are ranked and refined by the following RNN scoring-regression module. Samples are scored by accounting for accumulated future rewards, which enables better long-term strategic decisions similar to IOC frameworks. An RNN scene context fusion module jointly captures past motion histories, the semantic scene context and interactions among multiple agents. A feedback mechanism iterates over the ranking and refinement to further boost the prediction accuracy. We evaluate our model on two publicly available datasets: KITTI and Stanford Drone Dataset. Our experiments show that the proposed model significantly improves the prediction accuracy compared to other baseline methods.

%% file: introduction.tex
\section{Introduction}
\label{sec:intro}
\vspace{-2mm}
\epigraph{It is far better to foresee even without certainty than not to foresee at all. }{\textit{Henri Poincar\'{e} (Foundations of Science)}\vspace{-4mm}}



\begin{figure}
    \vspace{-4mm}
\centering
\begin{tabular}{l}
\scriptsize
\includegraphics[width=0.648\linewidth,trim=50mm 220mm 150mm 78mm,clip]{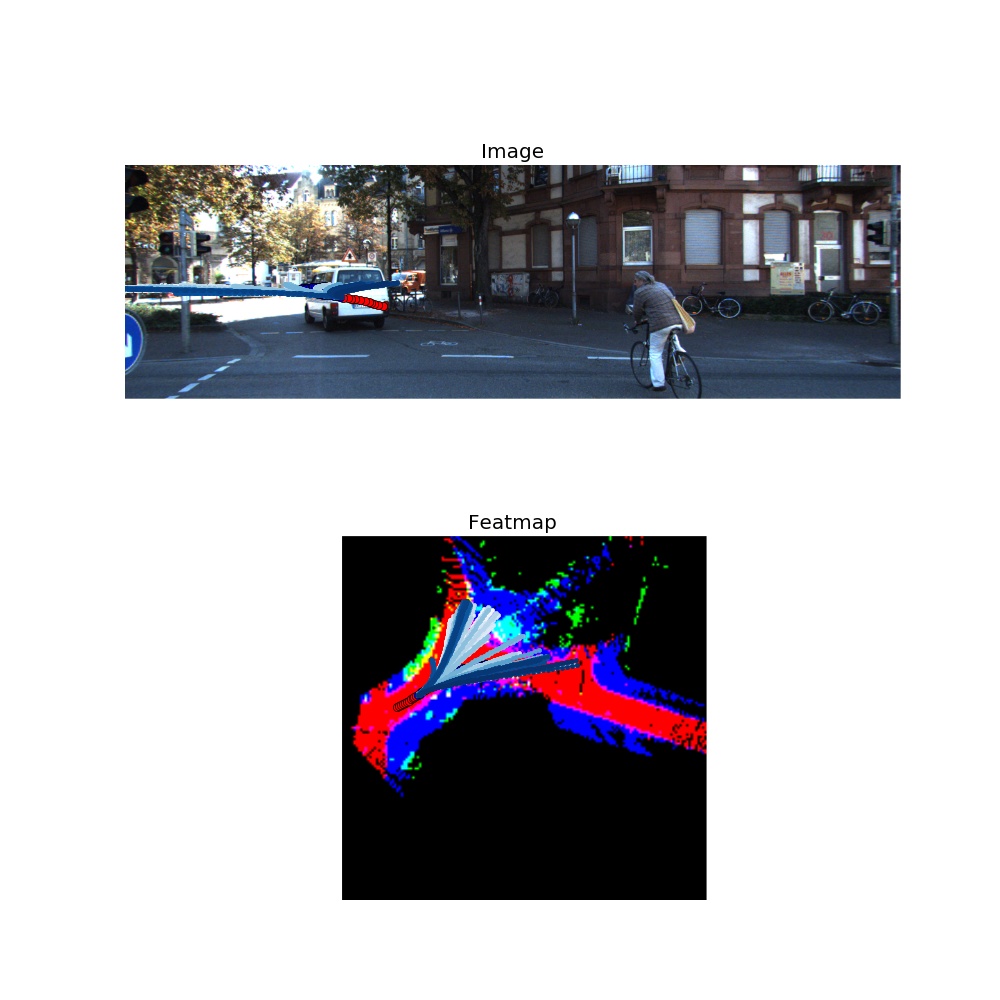}\includegraphics[width=0.33\linewidth,trim=130mm 84mm 125mm 200mm,clip]{figures/examples/kitti/intro_5_10.jpg}\\
\qquad\qquad\qquad (a) Future prediction example \\
\includegraphics[width=0.975\linewidth,trim=19mm 244mm 93mm 23mm,clip]{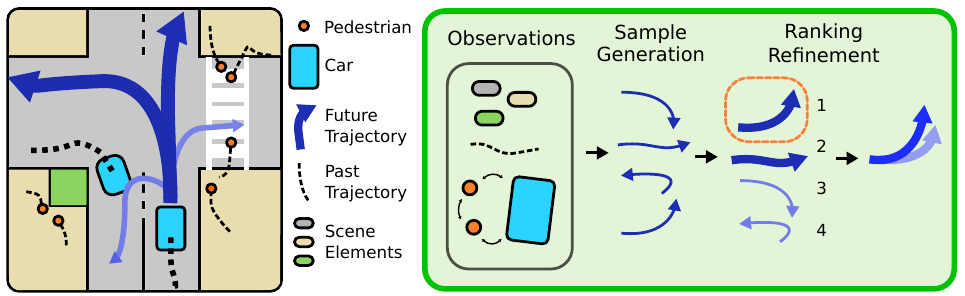}\\
\qquad\qquad\qquad (b) Workflow of \textit{DESIRE} \\
\end{tabular}
\caption{(a) A driving scenario: The white van may steer into left or right while trying to avoid a collision to other dynamic agents. \textit{DESIRE} produces accurate future predictions (shown as \textcolor{blue}{blue} paths) by tackling multi-modaility of future prediction while accounting for a rich set of both static and dynamic scene contexts. (b) \textit{DESIRE} generates a diverse set of hypothetical prediction samples, and then ranks and refines them through a deep IOC network.
}
\label{fig:intro}
\vspace{-3mm}
\end{figure}

Considering the future as a consequence of a series of past events, a \textit{prediction} entails reasoning about probable outcomes based on past observations. But predicting the future in many computer vision tasks is inherently riddled with uncertainty (see Fig.~\ref{fig:intro}). Imagine a busy traffic intersection, where such ambiguity is exacerbated by diverse interactions of automobiles, pedestrians and cyclists with each other, as well as with semantic elements such as lanes, crosswalks and traffic lights. Despite tremendous recent interest in future prediction \cite{alahisocial,ballan2016knowledge,jain2016recurrent,kitani2012activity,lee2016predicting,vondrickanticipating,walker2016uncertain}, existing state-of-the-art produces outcomes that are either deterministic, or do not fully account for interactions, semantic context or long-term future rewards.

In contrast, we present DESIRE, a {\underline{De}}ep {\underline{S}}tochastic {\underline{I}}OC {\underline{R}}NN {\underline{E}}ncoder-decoder framework, to overcome those limitations. The key traits of DESIRE are its ability to simultaneously: (a) generate {\em diverse hypotheses} to reflect a distribution over plausible futures, (b) reason about {\em interactions} between multiple dynamic objects and the scene context, (c) rank and refine hypotheses with consideration of {\em long-term future rewards}  (see Fig.~\ref{fig:intro}). These objectives are cast within a deep learning framework.

We model the scene as composed of semantic elements (such as roads and crosswalks) and dynamic participants or agents (such as cars and pedestrians). A static or moving observer is also considered as an instance of an agent. We formulate future prediction as determining the locations of agents at various instants in the future, relying solely on observations of the past states of the scene, in the form of agent trajectories and scene context derived from image-based features or other sensory data if available. The problem is posed in an optimization framework that maximizes the potential future reward of the prediction. Specifically, we propose the following novel mechanisms to realize the above advantages, also illustrated in Fig.~\ref{fig:overview_allinone}:

\vspace{-0.2cm}
\begin{tight_itemize}
\item \textit{Diverse Sample Generation}:
Sec.~\ref{sec:sampler} presents a conditional variational auto-encoder (CVAE) framework~\cite{sohn2015learning} to learn a sampling model that, given observations of past trajectories, produces a diverse set of prediction hypotheses to capture the multimodality of the space of plausible futures. The CVAE introduces a latent variable to account for the ambiguity of the future, which is combined with a recurrent neural network (RNN) encoding of past trajectories, to generate hypotheses using another RNN.

\item \textit{IOC-based Ranking and Refinement}:
    In Sec.~\ref{sec:ioc}, we propose a ranking module that determines the most likely hypotheses, while incorporating scene context and interactions. Since an optimal policy is hard to determine where multiple agents make strategic inter-dependent choices, the ranking objective is formulated to account for potential future rewards similar to inverse optimal control (IOC). This also ensures generalization to new situations further in the future, given limited training data. The module is trained in a multitask framework with a regression-based refinement of the predicted samples. In the testing phase, we iterate the above multiple times to obtain more accurate refinements of the future prediction. 
\item \textit{Scene Context Fusion}: 
    Sec.~\ref{sec:ssp} presents the Scene Context Fusion (SCF) layer that aggregates interactions between agents and the scene context encoded by a convolutional neural network (CNN). The fused embedding is channeled to the aforementioned RNN scoring module and allows to produce the rewards based on the contextual information.   

\end{tight_itemize}

While DESIRE is a general framework that is applicable to any future prediction task, we demonstrate its utility in two applications -- traffic scene understanding for autonomous driving and behavior prediction in aerial surveillance. Sec.~\ref{sec:experiments} demonstrates outstanding accuracy for predicting the future locations of traffic participants in the KITTI raw dataset and pedestrians in the Stanford Drone dataset.

To summarize, this paper presents DESIRE, which is a deep learning based stochastic framework for time-profiled distant future prediction, with several attractive properties:
\vspace{-0.2cm}
\begin{tight_itemize}
\item \textit{Scalability}: The use of deep learning rather than hand-crafted features enables end-to-end training and easy incorporation of multiple cues arising from past motions, scene context and interactions between multiple agents.
\item \textit{Diversity}: The stochastic output of a deep generative model (CVAE) is combined with an RNN encoding of past observations to generate multiple prediction hypotheses that hallucinate ambiguities and multimodalities inherent in future prediction.
%
\item \textit{Accuracy}: The IOC-based framework accumulates long-term future rewards for sampled trajectories and the regression-based refinement module learns to estimate a deformation of the trajectory, enabling more accurate predictions further into the future.
\end{tight_itemize}

%% file: related.tex
\section{Related Works}


\noindent \textbf{Classical methods}
Path prediction problems have been studied extensively with different approaches such as Kalman filters~\cite{kalman1960new}, linear regressions~\cite{mccullagh1989generalized} to non-linear Gaussian Process regression models~\cite{williams1998prediction, quinonero2005unifying, rasmussen2006gaussian, wang2008gaussian}, autoregressive models~\cite{akaike1969fitting} and time-series analysis~\cite{priestley1981spectral}. Such predictions suffice for scenarios with few interactions between the agent and the scene or other agents (like a flight monitoring system). In contrast, we propose methods for more complex environments such as surveillance for a crowd of pedestrians or traffic intersections, where the locomotion of individual agents is severely influenced by the scene context (\eg, drivable road or building) and the other agents (\eg, people or cars try to avoid colliding with the other). 	
	
\noindent \textbf{IOC for path prediction}
Kitani~\etal. recover human preferences (\ie, reward function) to forecast plausible paths for a pedestrian in~\cite{kitani2012activity} using inverse optimal control (IOC), or inverse reinforcement learning (IRL)~\cite{abbeel2004apprenticeship, ziebart2008maximum}, while~\cite{lee2016predicting} adapt IOC and propose a dynamic reward function to address changes in environments for sequential path predictions. Combined with a deep neural network, deep IOC/IRL has been proposed to learn non-linear reward functions and showed promising results in robot control~\cite{finn2016guided} and driving~\cite{wulfmeier2016watch} tasks. However, one critical assumption made in IOC frameworks, which makes them hard to be applied to general path prediction tasks, is that the goal state or the destination of agent should be given a priori, whereby feasible paths must be found to the given destination from the planning or control point of view. A few approaches relaxed this assumption with so-called goal set~\cite{mainprice2016goal, dragan2011manipulation}, but these goals are still limited to a target task space. Furthermore, a recovered cost function using IOC is inherently static, thus it is not suitable for time-profiled prediction tasks. Finally, past approaches do not incorporate interaction between agents, which is often a key constraint to the motion of multiple agents. In contrast, our methods are designed for more natural scenarios where agent goals are open-ended, unknown or time-varying and where agents interact with each other while dynamically adapting in anticipation of future behaviors.  

\noindent \textbf{Future prediction}
Walker~\etal ~\cite{walker2014patch} propose a visual prediction framework with a data-driven unsupervised approach, but only on a static scene, while~\cite{ballan2016knowledge} learn scene-specific motion patterns and apply to novel scenes for motion prediction as a knowledge transfer. A method for future localization from egocentric perspective is also addressed successfully in~\cite{park2016egocentric}. But unlike our method, none of those can provide time-profiled predictions. Recently, a large dataset is collected in \cite{robicquet2016learning} to propose the concept of social sensitivity to improve forecasting models and the multi-target tracking task. However, their social force~\cite{helbing1995social} based model has limited navigation styles represented merely using parameters of distance-based Gaussians.

\noindent \textbf{Interactions}
When modeling the behavior of an agent, it should also be taken into account that the dynamics of an agent not only depend on its own, but also on the behavior of others. Predicting the dynamics of multiple objects is also studied in~\cite{kooij2014context, kretzschmar2014learning, alahisocial, pellegrini2009you}, to name a few. Recently, a novel pooling layer is presented by \cite{alahisocial}, where the hidden state of neighboring pedestrians are shared together to joinly reason across multiple people. Nonetheless, these models lack predictive capacity as they do not take into account scene context. In~\cite{kooij2014context}, a dynamic Bayesian network to capture situational awareness is proposed as a context cue for pedestrian path prediction, but the model is limited to orientations and distances of pedestrians to vehicles and the curbside. A large body of work in reinforcement learning, especially game theoretical generalizations of Markov Decision Processes (MDPs), addresses multi-agent cases such as minmax-Q learning~\cite{littman1994markov} and Nash-Q learning~\cite{hu2003nash}. However, as noted in~\cite{shalev2016long}, typically learning in multi-agent setting is inherently more complex than single agent setting~\cite{shoham2007if, shoham2008multiagent, busoniu2008comprehensive}. 

\noindent \textbf{RNNs for sequence prediction}
Recurrent neural networks (RNNs) are natural generalizations of feedforward neural networks to sequences~\cite{sutskever2014sequence} and have achieved remarkable results in speech recognition~\cite{graves2013speech}, machine translation~\cite{bahdanau2014neural, sutskever2014sequence, cho2014learning} and image captioning~\cite{karpathy2015deep, xu2015show, donahue2015long}. The power of RNNs for sequence-to-sequence modeling thus makes them a reasonable model of choice to learn to generate sequential future prediction outputs. Our approach is similar to~\cite{cho2014learning} in making use of the encoder-decoder structure to embed a hidden representation for encoding and decoding variable length inputs and outputs. We choose to use gated recurrent units (GRUs) over long short-term memory units (LSTMs)~\cite{hochreiter1997long} since the former is found to be simpler yet yields no degraded performance~\cite{chung2014empirical}. Despite the promise inherent in RNNs, however, only a few works have applied RNNs to behavior prediction tasks. Multiple LSTMs are used in \cite{alahisocial} to jointly predict human trajectories, but their model is limited to producing fixed-length trajectories, whereas our model can produce variable-length ones. A Fusion-RNN that combines information from sensory streams to anticipate a driver's maneuver is proposed in \cite{jain2016recurrent}, but again their model outputs deterministic and fixed-length predictions.	

\noindent \textbf{Deep generative models}
Our work is also related to deep generative models~\cite{salakhutdinov2009deep, rezende2014stochastic, thibodeau2014deep}, as we have a sample generation process that is built on a variational auto-encoder (VAE)~\cite{kingma2013auto} within the framework. Since our prediction model essentially performs posterior-based probabilistic inference where candidate samples are generated based on conditioning variables (\ie, past motions besides latent variables), we naturally extend our method to exploit a conditional variational auto-encoder (CVAE)~\cite{kingma2014semi, sohn2015learning} during the sample generation process. Dense trajectories of pixels are predicted from a single image using CVAE in \cite{walker2016uncertain}, while we focus on predicting long-term behaviors of multiple interacting agents in dynamic scenes. 

Unlike our framework, all aforementioned approaches lack either consideration of scene context, modeling of interaction with other agents or capabilities in producing continuous, time-profiled and long-term accurate predictions.

%% file: method.tex
\section{Method}
\label{sec:method}

\begin{figure*}[t]
\vspace{-8mm}
\centering
\includegraphics[width=0.95\linewidth,trim=20mm 205mm 12mm 20mm,clip]{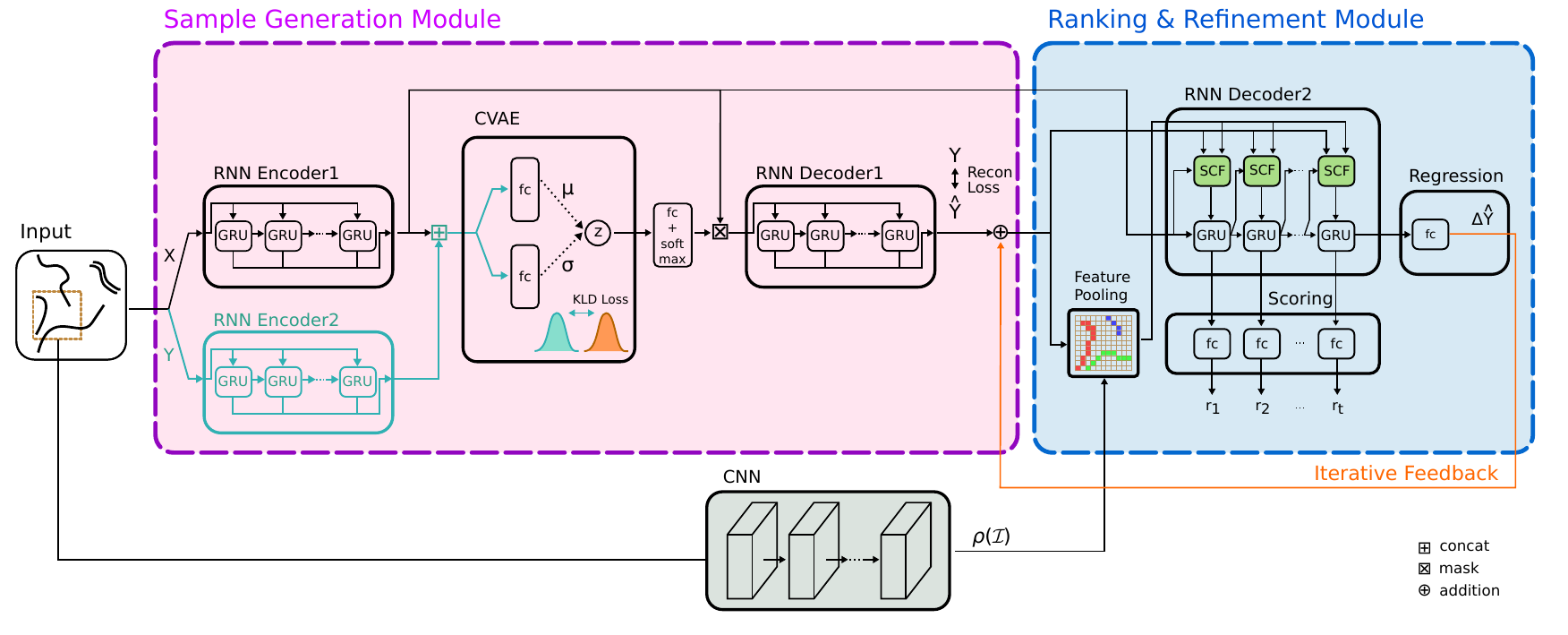}
\vspace{-3mm}
\caption{
The overview of proposed prediction framework \textit{DESIRE}. First, DESIRE generates multiple plausible prediction samples $\hat{Y}$ via a CVAE-based RNN encoder-decoder (\textcolor{Purple}{Sample Generation Module}). Then the following module assigns a reward to the prediction samples at each time-step sequentially as IOC frameworks and learns displacements vector $\Delta{\hat{Y}}$ to regress the prediction hypotheses (\textcolor{Blue}{Ranking and Refinement Module}). The regressed prediction samples are refined by iterative feedback. The final prediction is the sample with the maximum accumulated future reward. Note that the flow via \textcolor{Aquamarine}{aquamarine-colored paths} is only available during the training phase.}
\label{fig:overview_allinone}
\vspace{-6mm}
\end{figure*}

We formulate the future prediction problem as an optimization process, where the objective is to learn the posterior distribution $P(\mathbf{Y}|\mathbf{X}, \mathcal{I})$ of multiple agents' future trajectories $\mathbf{Y} = \{Y_1,Y_2,..,Y_n\}$ given their past trajectories $\mathbf{X}=\{X_1,X_2,..,X_n\}$ and sensory input $\mathcal{I}$ where $n$ is the number of agents. The future trajectory of an agent $i$ is defined as $Y_i = \{y_{i,t+1}, y_{i,t+2}, .., y_{i,t+\delta}\}$, and the past trajectory is defined similarly as $X_i = \{x_{i,t-\iota+1}, x_{i,t-\iota+2}, .., x_{i,t}\}$. Here, each element of a trajectory (\eg, $y_{i,t}$) is a vector in $\mathbb{R}^2$ (or $\mathbb{R}^3$) representing the coordinates of agent $i$ at time $t$, and $\delta$ and $\iota$ refer to the maximum length of time steps for future and past respectively.
Since direct optimization of continuous and high dimensional $\textbf{Y}$ is not feasible, we design our method to first sample a diverse set of future predictions and assign a probabilistic score to each of the samples to approximate $P(\mathbf{Y}|\mathbf{X}, \mathcal{I})$.   
In this section, we describe the details of DESIRE (Fig.~\ref{fig:overview_allinone}) in the following structure: \textit{Sample Generation Module} (Sec.~\ref{sec:sampler}), \textit{Ranking and Refinement Module} (Sec.~\ref{sec:ioc}), and \textit{Scene Context Fusion} (Sec.~\ref{sec:ssp}).

\subsection{Diverse Sample Generation with CVAE}
\label{sec:sampler}

Future prediction can be inherently ambiguous and has uncertainties as multiple plausible scenarios can be explained under the same past situation (\eg, a vehicle heading toward an intersection can make different turns as seen in Fig.~\ref{fig:intro}). Thus, learning a deterministic function $f$ that directly maps $\{\mathbf{X},\mathcal{I}\}$ to $\mathbf{Y}$
will under-represent potential prediction space and easily over-fit to training data. Moreover, a naively trained network with a simple loss will produce predictions that average out all possible outcomes.

In order to tackle the uncertainty, we adopt a deep generative model, conditional variational auto-encoder (CVAE)~\cite{sohn2015learning}, inside of DESIRE framework.
CVAE is a generative model that can learn the distribution $P(Y_i | X_i)$ of the output $Y_i$ conditioned on the input $X_i$ by introducing a stochastic latent variable $z_i$\footnote{Notice that we learn the distribution independently over different agents in this step. Interaction between agents is considered in Sec.~\ref{sec:ioc}.}.
It is composed of multiple neural networks, such as recognition network $Q_\phi(z_i|Y_i,X_i)$, (conditional) prior network $P_\nu(z_i|X_i)$, and generation network $P_\theta(Y_i|X_i,z_i)$. 
Here, $\theta, \phi, \nu$ denote the parameters of corresponding networks. 
The prior of the latent variables $z_i$ is modulated by the input $X_i$, however, this can be relaxed to make the latent variables statistically independent of input variables, \ie, $P_\nu(z_i|X_i) = P_\nu(z_i)$~\cite{kingma2014semi,sohn2015learning}. Essentially, a CVAE introduces stochastic latent variables $z_i$ that are learned to encode a diverse set of predictions $Y_i$ given input $X_i$, making it suitable for modeling one-to-many mapping. During training, $Q_\phi(z_i|Y_i,X_i)$ is learned such that it gives higher probability to $z_i$ that is likely to produce a reconstruction $\hat{Y_i}$ close to actual prediction given the full context $X_i$ and $Y_i$. At test time $z_i$ is sampled randomly from the prior distribution and decoded through the decoder network to produce a prediction hypothesis. This enables probabilistic inference which serves to handle multi-modalities in the prediction space. 

\noindent\textbf{Train phase}:
Firstly, the past and future trajectories of an agent $i$, $X_i$ and $Y_i$ respectively, are encoded through two RNN encoders with separate set of parameters (\ie, RNN Encoder1 and RNN Encoder2 in Fig.~\ref{fig:overview_allinone}). The resulting two encodings, $\mathcal{H}_{X_i}$ and $\mathcal{H}_{Y_i}$, are concatenated and passed through one fully connected ($fc$) layer with a non-linear activation (\eg, $relu$). Two side-by-side $fc$ layers are followed to produce both the mean $\mu_{z_i}$ and the standard deviation $\sigma_{z_i}$ over $z_i$. The distribution of $z_i$ is modeled as a Gaussian distribution (\ie, $z_i \sim Q_\phi(z_i|X_i,Y_i) = \mathcal{N}(\mu_{z_i}, \sigma_{z_i})$) and is regularized by the $\mathcal{KL}$ divergence against a prior distribution $P_\nu(z_i) := \mathcal{N}(0,I)$ during the training. Upon successful training, the target distribution is learned in the latent variable $z_i$, which allows one to draw a random sample $z_i$ from a Gaussian distribution to reconstruct $Y_i$ at test time. 
Since back-propagation is not possible through random sampling, we adopt the standard \textit{reparameterization trick}~\cite{kingma2013auto} to make it differentiable. 

In order to model $P_\theta(Y_i|X_i,z_i)$, $z_i$ is combined with $X_i$ as follows. The sampled latent variable $z_i$ is passed to one $fc$ layer to match the dimension of $\mathcal{H}_{X_i}$ that is followed by a $softmax$ layer, producing $\beta(z_i)$. Then that is combined with the encodings of past trajectories $\mathcal{H}_{X_i}$ through a masking operation $\boxtimes$ (\ie, element-wise multiplication). One can interpret this as a \textit{guided drop out} where the guidance $\beta$ is derived from the full context of individual trajectory during the training phase, while it is randomly drawn from ${X_i}, {Y_i}$ agnostic prior distribution $z_i^{(k)} \sim P_\nu(z_i)$ in the testing phase. Finally, the following RNN decoder (\ie, RNN Decoder1 in Fig.~\ref{fig:overview_allinone}) takes the output of the previous step, $\mathcal{H}_{X_i} \boxtimes \beta({z_i^{(k)}})$, and generates $K$ number of future prediction samples, \ie, $\hat{{Y_i}}^{(1)}, \hat{{Y_i}}^{(2)}, .., \hat{{Y_i}}^{(K)}$. 

There are two loss terms in training the CVAE-based RNN encoder-decoder. 
\vspace{-2mm}
\begin{tight_itemize}
    \item Reconstruction Loss: $\ell_{Recon} = \frac{1}{K} \sum_{k} \Vert {Y_i} - \hat{{Y_i}}^{(k)} \Vert$. This loss measures how far the generated samples are from the actual ground truth. 
    \item $\mathcal{KLD}$ Loss: $\ell_{KLD} = D_{\mathcal{KL}}(Q_\phi({z_i} | {Y_i},{X_i})\Vert P_\nu({z_i}))$. This regularization loss measures how close the sampling distribution at test time is to the distribution of latent variable that we learn during training. 
\end{tight_itemize}

\noindent\textbf{Test phase}:
At test time, the encodings of future trajectories $\mathcal{H}_{Y_i}$ are not available, thus the encodings of past trajectories $\mathcal{H}_{X_i}$ are combined with multiple random samples of latent variable ${z_i^{(k)}}$ drawn from the prior $z_i^{(k)} \sim P_\nu({z_i})$. Similar to the training phase, $\mathcal{H}_{X_i} \boxtimes \beta({z_i^{(k)}})$ is passed to the following RNN decoder (\ie, RNN Decoder1 in Fig.~\ref{fig:overview_allinone}) to generate a diverse set of prediction hypotheses. 

\vspace{1mm}
\noindent\textbf{Further details}: For both train and test phases, we pass trajectories through a temporal convolution layer before encoding to encourage the network to learn the concept of velocity from adjacent frames before getting passed into RNN encoders. Also, RNNs are implemented using gated recurrent units (GRU)~\cite{cho2014learning} to learn long-term dependencies, yet they can be easily replaced with other popular RNNs like long short-term memory units (LSTM)~\cite{hochreiter1997long}. In summary, this sample generation module produces a set of diverse hypotheses critical to capturing the multimodality of the prediction task, through a effective combination of CVAE and RNN encoder-decoder. Unlike~\cite{walker2016uncertain}, where CVAE is used to predict for short-term visual motion from a single image, our CVAE module generates diverse set of future trajectories based on a past trajectory. 

\subsection{IOC-based Ranking and Refinement}
\label{sec:ioc}

Predicting a \textit{distant} future can be far more challenging than predicting one close by. In order to tackle this, we adopt the concept of decision-making process in reinforcement learning (RL) where an agent is trained to choose its actions that maximizes \textit{long-term rewards} to achieve its goal~\cite{sutton1998reinforcement}. Instead of designing a reward function manually, however, IOC~\cite{wulfmeier2016watch, finn2016guided} learns an unknown reward function. 
Inspired by this, we design an RNN model that assigns rewards to each prediction hypothesis $\hat{Y_i}^{(k)}$ and measures their \textit{goodness} $s_i^{(k)}$ based on the accumulated long-term rewards. Thereafter, we also directly refine prediction hypotheses by learning displacements $\triangle \hat{Y_i}^{(k)}$ to the actual prediction through another $fc$ layer. Lastly, the module receives iterative feedbacks from regressed predictions and keeps adjusting so that it produces precise predictions at the end. The model is illustrated in the right side of Fig.~\ref{fig:overview_allinone}. During the process, we combine 1) past motion history through the embedding vector $\mathcal{H}_\mathbf{X}$, 2) semantic scene context through a CNN with parameters $\rho$, and 3) interaction among multiple agents by using interaction features (Sec.~\ref{sec:ssp}). Notice that unlike typical robotics applications~\cite{wulfmeier2016watch, finn2016guided}, we do not assume that the goal (final destination) is known or the dynamics of the agents are given. Our model learns the agents dynamics as well as the scene context in a coherent framework.

\vspace{1mm}
\noindent\textbf{Learning to score}: For an agent $i$, there are $K$ number of samples (\ie, $\hat{Y}_i^{(1)},\hat{Y}_i^{(2)}, .., \hat{Y}_i^{(K)}$) that are generated by our CVAE sampler. Let the score $s$ of individual prediction hypothesis $\hat{Y}_i^{(k)}$ for the agent $i$ be defined as follows, 
\vspace{-4mm}
\begin{equation}
    s(\hat{Y}_i^{(k)}; \mathcal{I}, \mathbf{X}, \hat{\mathbf{Y}}_{j \backslash i}^{(\forall)}) 
     = \sum_{t=1}^{T} \psi(\hat{y}_{i,t}^{(k)}; \mathcal{I}, \mathbf{X}, \hat{\mathbf{Y}}_{\tau < t}^{(\forall)}),
\label{eq:ioc}
\end{equation}
where $\hat{\mathbf{Y}}_{j \backslash i}^{(\forall)}$ is the prediction samples of other agents (\ie, $\forall j$, where $j \neq i$),  $\hat{y}_{i,t}^{(k)}$ is the $k^{th}$ prediction sample of an agent $i$ at time $t$, $\hat{\mathbf{Y}}_{\tau < t}^{(\forall)}$ is all the prediction samples until a time-step $t$, $T$ is the maximum prediction length, and $\psi$ is the reward function that assigns a reward value at each time-step. $\psi$ is implemented as an $fc$ layer that is connected to the hidden vector of RNN cell at each time step. We share the parameters of the $fc$ layer over all the time steps (each RNN cell outputs the hidden state of the same dimension).
 Therefore, the score $s$ is accumulated rewards over time, accounting for the entire future rewards being assigned to each hypothesis. This enables our model to make a strategic decision by allowing us to rank samples as in other sampling-based IOC frameworks~\cite{finn2016guided}. In addition, the reward function $\psi$ incorporates both scene context $\mathcal{I}$ as well as the interaction between agents (see Sec.~\ref{sec:ssp}).

\vspace{1mm}
\noindent\textbf{Learning to refine}: Alongside the scores, our model also estimates a regression vector $\triangle\hat{Y}_i^{(k)}$ that refines each prediction sample $\hat{Y}_i^{(k)}$. The regression vector for each agent $i$ is obtained with the regression function $\eta$ defined as follows, 
\vspace{-2mm}
\begin{equation}
    \triangle\hat{Y}_i^{(k)} = \eta(\hat{Y}_i^{(k)}; \mathcal{I}, \mathbf{X}, \hat{\mathbf{Y}}_{j \backslash i}^{(\forall)}).
\label{eq:reg}	
\vspace{-2mm}
\end{equation}
Represented as parameters of a neural network, the regression function $\eta$ accumulates both scene contexts and all other agents dynamics from the past to entire future frames, and estimates the best displacement vector $\triangle\hat{Y}_i^{(k)}$ over entire time-horizon $T$. Similarly to the score $s$, it accounts for what happens in the future both in terms of scene context and interactions among dynamic agents to produce the output. We implement $\eta$ as another $fc$ layer that is connected to the last hidden vector of the RNN which outputs $M \times T$ dimensional vector. $M=2$ (or $3$) is the dimension of the location state. 

\noindent\textbf{Iterative feedback}: Using the displacement vector $\triangle\hat{Y}_i^{(k)}$, we iteratively refine the prediction hypothesis $\hat{Y}_i^{(k)}$. After each cycle, $\hat{Y}_i^{(k)}$ is updated by $\hat{Y}_i^{(k)} + \triangle\hat{Y}_i^{(k)}$, and fed into the IOC module. This process is similar to the gradient descent optimization of $\hat{Y}_i$ over the score function $s$, but it does not require to compute the gradient over RNN which can be very unstable due to the recurrent structure (i.e., vanishing or exploding gradient). We observe that iterative refinement indeed improves the quality of prediction samples in the experiments (see Fig.~\ref{fig:quant} and Fig.~\ref{fig:iter}).

\noindent\textbf{Losses}: There are two loss terms in training the IOC ranking and refinement module.
\vspace{-2mm}
\begin{tight_itemize}
    \item Cross-entropy Loss: $\ell_{CE} = H(p,q)$ of which the target distribution $q$ is obtained by $softmax(-d(Y_i, \hat{Y}_i^{(k)}))$, where $d(Y_i, \hat{Y}_i^{(k)}) = \max \Vert\hat{Y}_i^{(k)} - Y_i \Vert$. 
    \item Regression Loss: $\ell_{Reg} = \frac{1}{K} \sum_k \Vert {Y_i} - \hat{{Y_i}}^{(k)} - \triangle\hat{{Y_i}}^{(k)} \Vert$ 
\end{tight_itemize}
Finally, the total loss of the entire network is defined as a multi-task loss as follows, where $N$ is the number of agents in one batch.
\vspace{-2mm}
\begin{equation}
    \ell_{Total} = \frac{1}{N} \sum_{i \in N} \ell_{Recon} +  \ell_{KLD} +  \ell_{CE} + \ell_{Reg}
\label{eq:totalloss}	
\end{equation}

\subsection{Scene Context Fusion}
\label{sec:ssp}

\begin{figure}
\vspace{-4mm}
\centering
\includegraphics[width=0.8\linewidth,trim=20mm 205mm 85mm 15mm,clip]{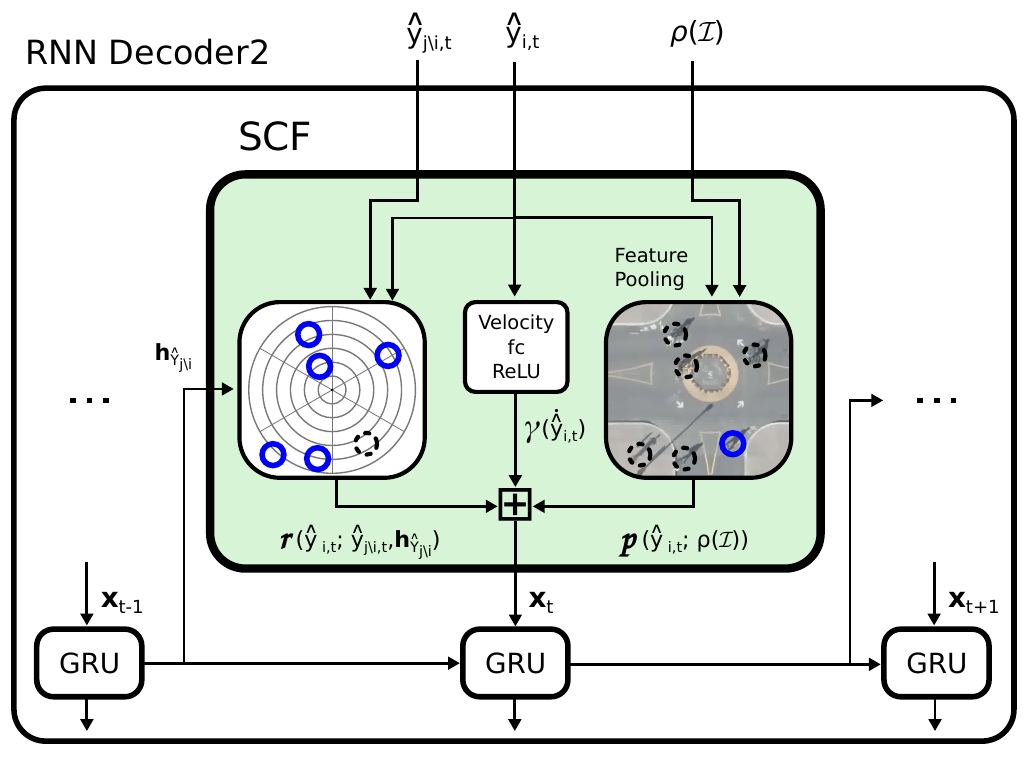}
\caption{
    Details of Scene Context Fusion unit (SCF) in RNN Decoder2 in Fig.~\ref{fig:overview_allinone}. Note that the input to the GRU cell at each time-step, $\mathbf{x}_t$, integrates multiple cues (\ie, the dynamics of agents, scene context and interaction between agents). 
}
\label{fig:RNNDecoder2_details}
\vspace{-5mm}
\end{figure}

As discussed in the previous section, our ranking and refinement module relies on the hidden representation of the shared RNN module. Thus, it is important that the RNN must contain the information about 1) individual past motion context, 2) semantic scene context and 3) the interaction between multiple agents, in order to provide proper hidden representations that can score and refine a prediction $\hat{Y}_i^{(k)}$. 

We achieve the goal by having an RNN that takes following input $\mathbf{x}_t$ at each time step:
\begin{equation}
    \mathbf{x}_t = \Big[\gamma(\hat{v}_{i,t}), p(\hat{y}_{i,t}; \rho(\mathcal{I})), r(\hat{y}_{i,t}; \hat{y}_{j \backslash i,t}, \mathbf{h}_{\hat{Y}_{j \backslash i}})\Big]
    \label{eq:xin}
    \vspace{-2mm}
\end{equation}
where $\hat{v}_{i,t}$ is a velocity of $\hat{Y}_i^{(k)}$ at $t$, $\gamma$ is a $fc$ layer with a \emph{ReLU} activation that maps the velocity to a high dimensional representation space, 
$p(\hat{y}_{i,t}; \rho(\mathcal{I}))$ is a pooling operation that pools the CNN feature $\rho(\mathcal{I})$ at the location $\hat{y}_{i,t}$, 
$r(\hat{y}_{i,t}; \hat{y}_{j \backslash i, t}, \mathbf{h}_{\hat{Y}_{j \backslash i}})$ is the interaction feature computed by a fusion layer that spatially aggregates other agents hidden vectors, similar to SocialPooling (SP) layer~\cite{alahisocial}. The embedding vector $\mathcal{H}_{X_i}$ (the output of the RNN Encoder1 in Fig.~\ref{fig:overview_allinone}) is shared as the initial hidden state of the RNN, in order to provide the individual past motion context. We share this embedding with the CVAE module since both require the same information to be embedded in the vector.

\noindent \textbf{Interaction Feature}: We implement a spatial grid based pooling layer similar to SP layer~\cite{alahisocial}. For each sample $k$ of an agent $i$ at $t$, we define spatial grid cells centered at $\hat{y}_{i,t}^{(k)}$. Over each grid cell $g$, we pool the hidden representation of all the other agents' samples that are within the spatial cell, $\forall j \neq i, \forall k, \hat{y}_{j, t}^{(k)} \in g$. Instead of using the max pooling operation with rectangular grids, we adopt log-polar grids with an average pooling. Combined with CNN features, the SCF module provides the RNN decoder with both static and dynamic scene information. It learns consistency between semantics of agents and scenes for reliable prediction.

\subsection{Characteristics of DESIRE}
\label{sec:highlights}

This section highlights particularly distinctive features of DESIRE that naturally enable higher accuracy and reliability.
\vspace{-0.6cm}
\begin{tight_itemize}
    \item The framework is based on deep neural network and is trainable end-to-end, rather than relying on hand-crafted parametric representation and interactions terms. Trajectories of each agent are represented using RNN encoders and are combined together through a fusion layer within the architecture. Scene context is represented through CNN and is not solely restricted to images (\ie, can handle non-visual sensors too). Overall, the algorithm is scalable and flexible.
    \item CVAE is combined with RNN encodings to generate stochastic prediction hypothesis, which handles ambiguities and multimodalities inherent in future prediction.
    \item A novel RNN module coherently integrates multiple cues that have critical influence on behavior prediction such as dynamics of all neighboring agents and scene semantics.
    \item An IOC framework is used to train the trajectory ranking objective by measuring potential long-term future rewards. This makes the model less reactive, and enables more accurate predictions further into the future.
    \item A regression vector is learned to refine trajectories and an iterative feedback mechanism sequentially adjusts the predicted behavior, resulting in more accurate predictions.
\end{tight_itemize}

%% file: experiments.tex
\section{Experiments} \label{sec:experiments} 

\subsection{Datasets}
\label{sec:datasets}
\noindent \textbf{KITTI Raw Data}~\cite{Geiger2013IJRR}: The dataset provides images of driving scenes and Velodyne 3D laser scan along with calibration information between cameras and sensors. To prepare data examples (\ie, $X, Y, \mathcal{I}$), we performed the following: As the dataset does not provide semantic labels for 3D points (which we need for \textit{scene context}), we first perform semantic segmentations of images and project Velodyne laser scans onto the image plane using the provided camera matrix to label 3D points. The semantically labeled 3D points are then registered into the world coordinates using GPS-IMU tags. Finally we create top-down view feature maps $\mathcal{I}$ of size $H \times W \times C$ ($H, W$: size of crop and $C$: number of classes for scene elements, \eg, road, sidewalk, and vegitation shown as red, blue and green color in Fig.~\ref{fig:kitti_qual}.). $\mathcal{I}$ is cropped with respect to the view point of the camera to simulate actual driving scenario ($H,W=80m$ and the size of pixel is $0.5m$. The camera is located at the left-center.). 
Since laser scans on dynamic objects generate traces during registration, we remove moving objects and only use static scene elements. The trajectories $X,Y$ are generated by extracting the center locations of the 3D tracklets and registering them in the world coordinates. We use all annotated videos from Road and City scenes for our experiments and generate approximately 2,500 training examples.

\noindent \textbf{Stanford Drone Dataset}~\cite{robicquet2016learning}: The dataset contains a large volume of aerial videos captured in a university campus using a drone. There are various classes of dynamic objects interacting with each other, often in the form of high density crowds. Except for less stabilized cameras and lost labels, we used all videos to create examples to train/test our model, yielding approximately $16,000$ examples. Note that we directly use raw images to extract visual features, rather than semantically labeled feature maps. We resize the images by $1/5$ in following experiments to avoid memory overhead. 

\subsection{Evaluation Metrics and Baselines}
\label{sec:metrics}

\begin{figure}
    \vspace{-8mm}
\centering
\begin{tabular}{@{\hspace{0mm}}c@{\hspace{0mm}}c@{\hspace{0mm}}c@{\hspace{0mm}}c@{\hspace{0mm}}}
\includegraphics[width=0.28\linewidth,trim=8mm 0mm 20mm 7mm,clip]{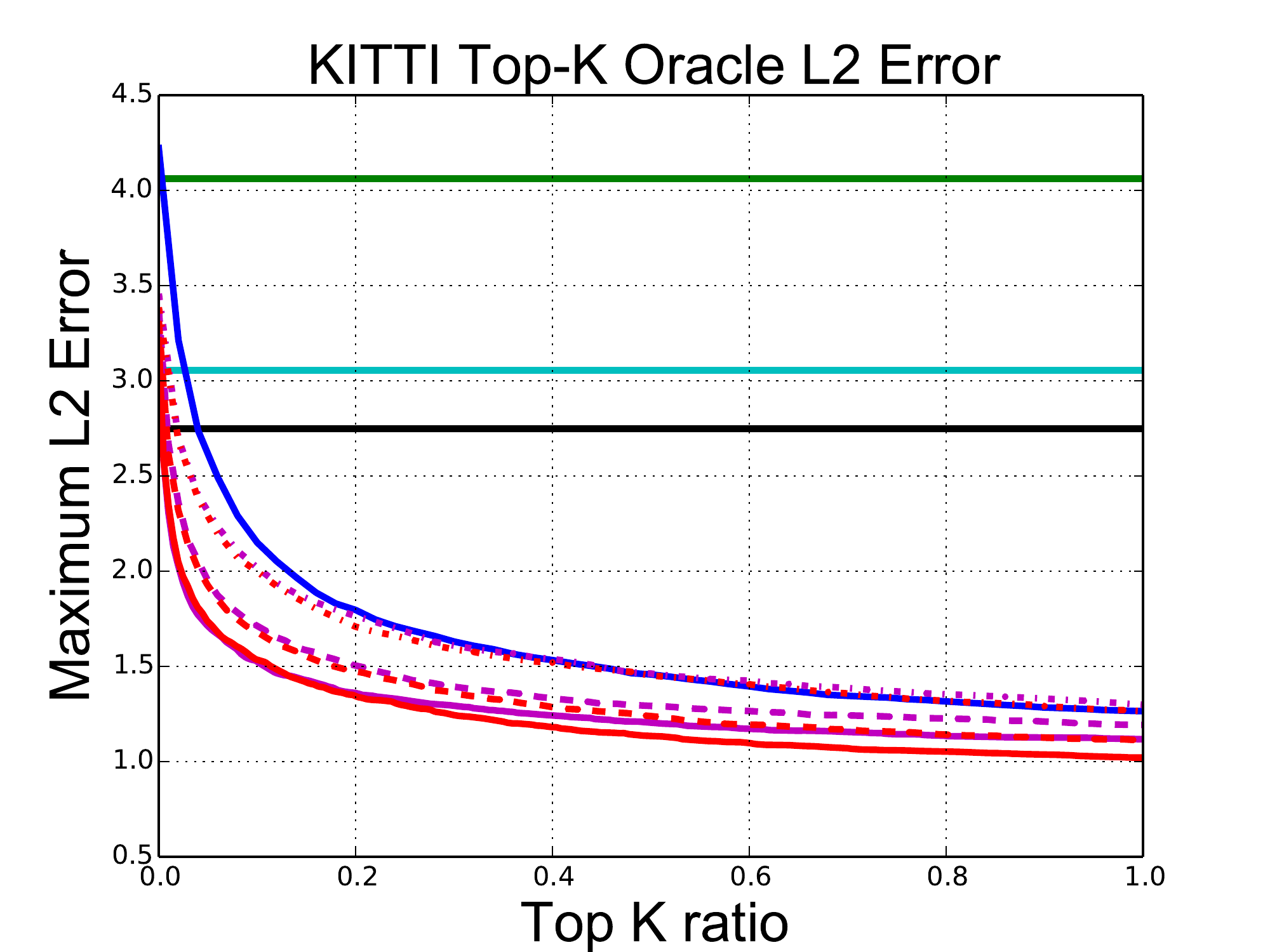} &
\includegraphics[width=0.28\linewidth,trim=8mm 0mm 20mm 7mm,clip]{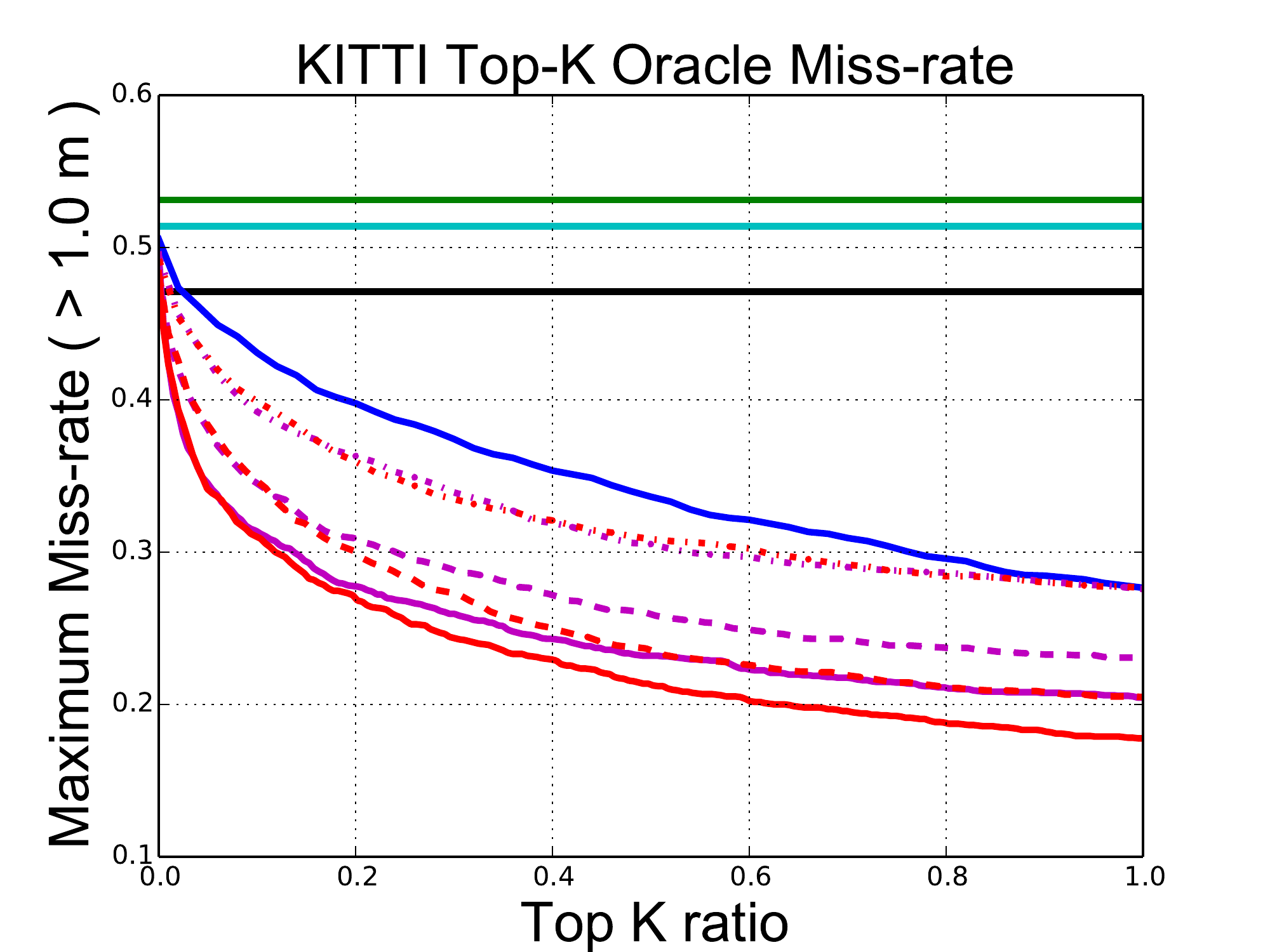}&
\includegraphics[width=0.28\linewidth,trim=8mm 0mm 20mm 7mm,clip]{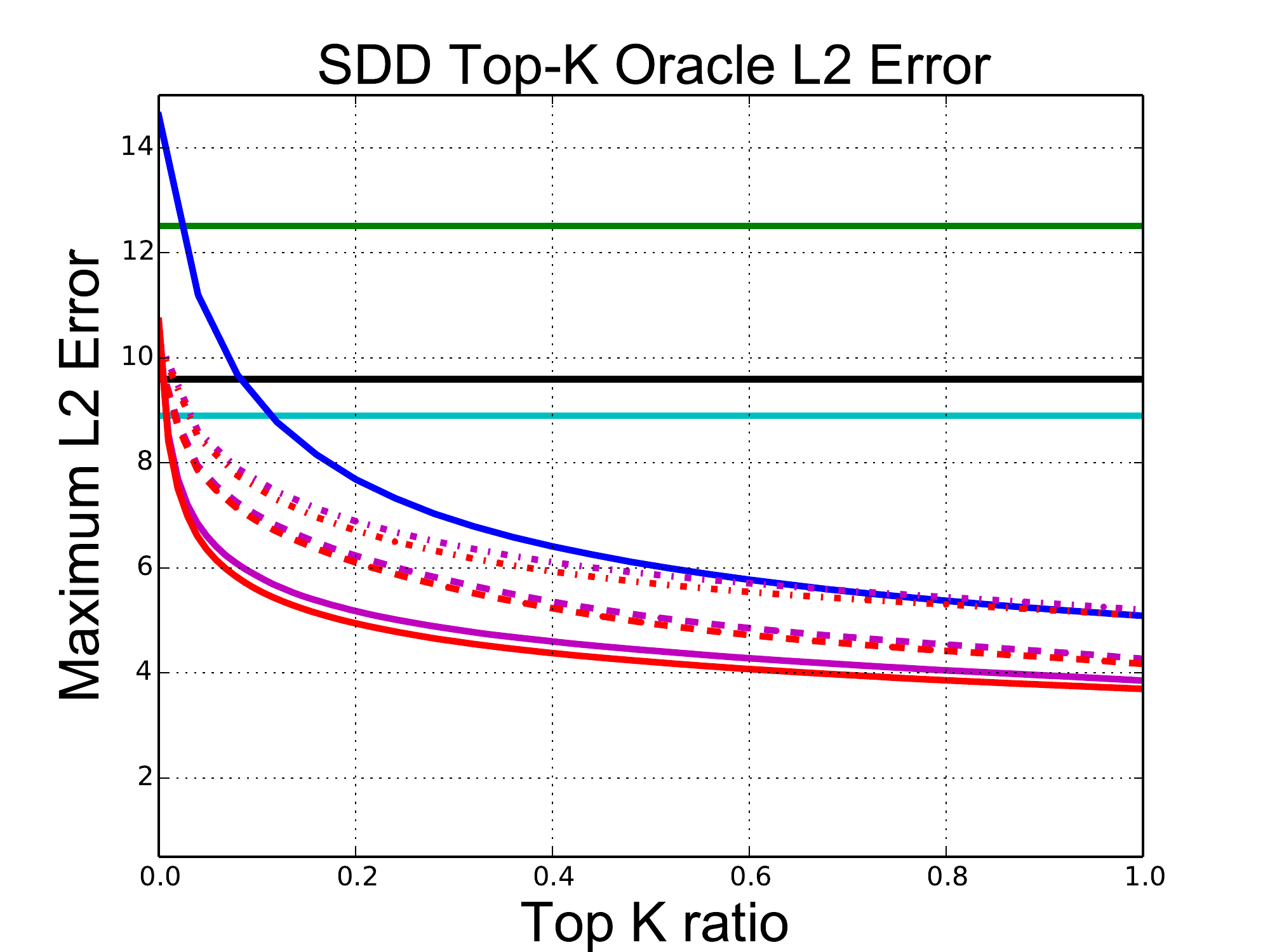} & 
\includegraphics[width=0.146\linewidth,trim=140mm 72mm 21.5mm 17mm,clip]{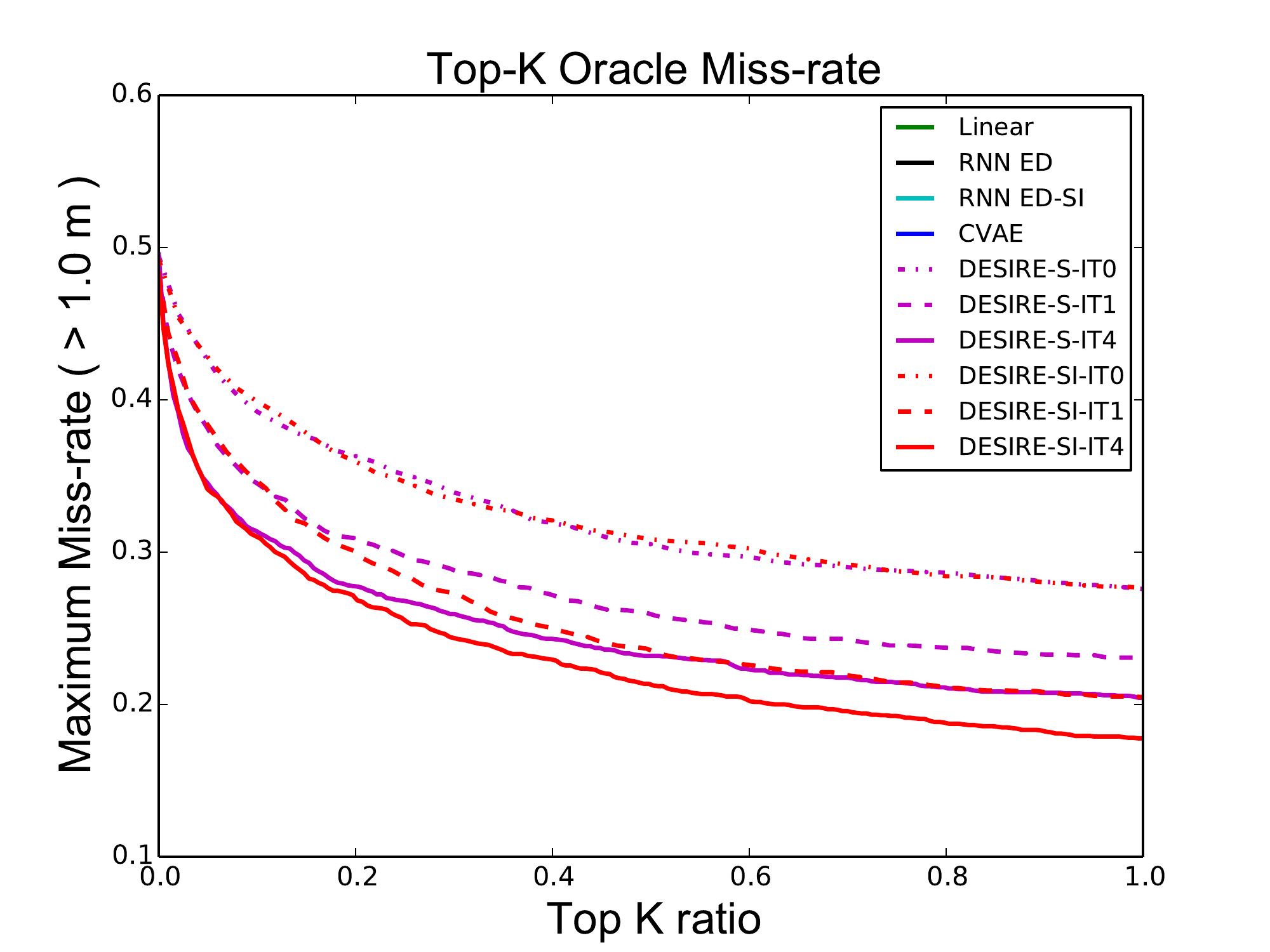}\\
\end{tabular}
\caption{Oracle prediction errors over the number of samples on the KITTI dataset. X axis represents the ratio of top samples used in the oracle error evaluation (Y axis). Best viewed in color.
}
\label{fig:quant}
\vspace{-4mm}
\end{figure}

The following metrics are used to measure the performance of future prediction task in various aspects: (i) L2 distance between the prediction and ground truth at multiple time steps, (ii) miss-rate with a threshold in terms of L2 distance at multiple time steps, (iii) maximum L2 distance over entire time frames, (iv) maximum miss-rate over entire time frames, and (v) \textit{oracle} error over top K number of samples (\ie, $\mathcal{E}_{oracle} = \min_{\forall k \in K} \mathcal{E}(\hat{Y}_i^{(k)} - Y_i)$) to account for the uncertainty in the future prediction (similar to MEE in~\cite{walker2016uncertain}). We set K to be $50$ throughout the main experiments.

We compare our method with the following baselines:
\vspace{-2mm}
\begin{tight_itemize}
    \item \textit{Linear}: A linear regressor that estimates linear parameters by minimizing the least square error.
    \item \textit{RNN ED}: An RNN encoder-decoder model that directly regresses the prediction only using the past trajectories. 
    \item \textit{RNN ED-SI}: An \textit{RNN ED} augmented with our SCF unit into the decoder similar to~\cite{jain2016recurrent}. The model combines the scene and interaction features while making prediction and uses the same information as ours, but makes a prediction at $t+1$ solely based on the past information up to $t$. 
    \item \textit{DESIRE}: The proposed method. We denote our model with only semantic scene context in SCF module as \textit{DESIRE-S} and our model with both scene context and interaction as \textit{DESIRE-SI}. We also evaluate \textit{DESIRE-X-IT\{N\}}, where N is the number of iterative feedbacks.
\end{tight_itemize}

\subsection{Learning Details}
\label{sec:learningdetails}
We train the model with Adam optimizer~\cite{kingma2014adam} with the initial learning rate of $0.004$. The learning rate is decreased by half at every quarter of total epochs, albeit we do not observe clear improvement with this. 
All the models including Encoder-decoder baselines are trained for $600$ epochs for KITTI and $8$ epochs for SDD (about $50$K iterations with a batch size $32$). The full details on the architecture are discussed in the supplementary materials.
In order to avoid exploding gradient in RNNs, we apply gradient clipping with L2 norm of $1.0$. 
During the training procedure, we randomly rotate the scene and trajectories to augment data and reduce over-fitting. For all experiments, we run randomized 5 fold cross validation without overlapping videos in different splits. All models observe maximum of $2$ seconds for past trajectories and make a prediction up to $4$ seconds into the future. 
All models are implemented using TensorFlow and trained end-to-end with a NVIDIA Tesla K80 GPU. Training takes approximately one to two days per model. 

\subsection{Analysis}
\label{sec:quantitative}

\begin{figure}
    \vspace{-8mm}
\centering
\begin{tabular}{@{\hspace{0mm}}c@{\hspace{0mm}}c@{\hspace{0mm}}c@{\hspace{0mm}}}
\includegraphics[width=0.33\linewidth,trim=80mm 210mm 100mm 58mm,clip]{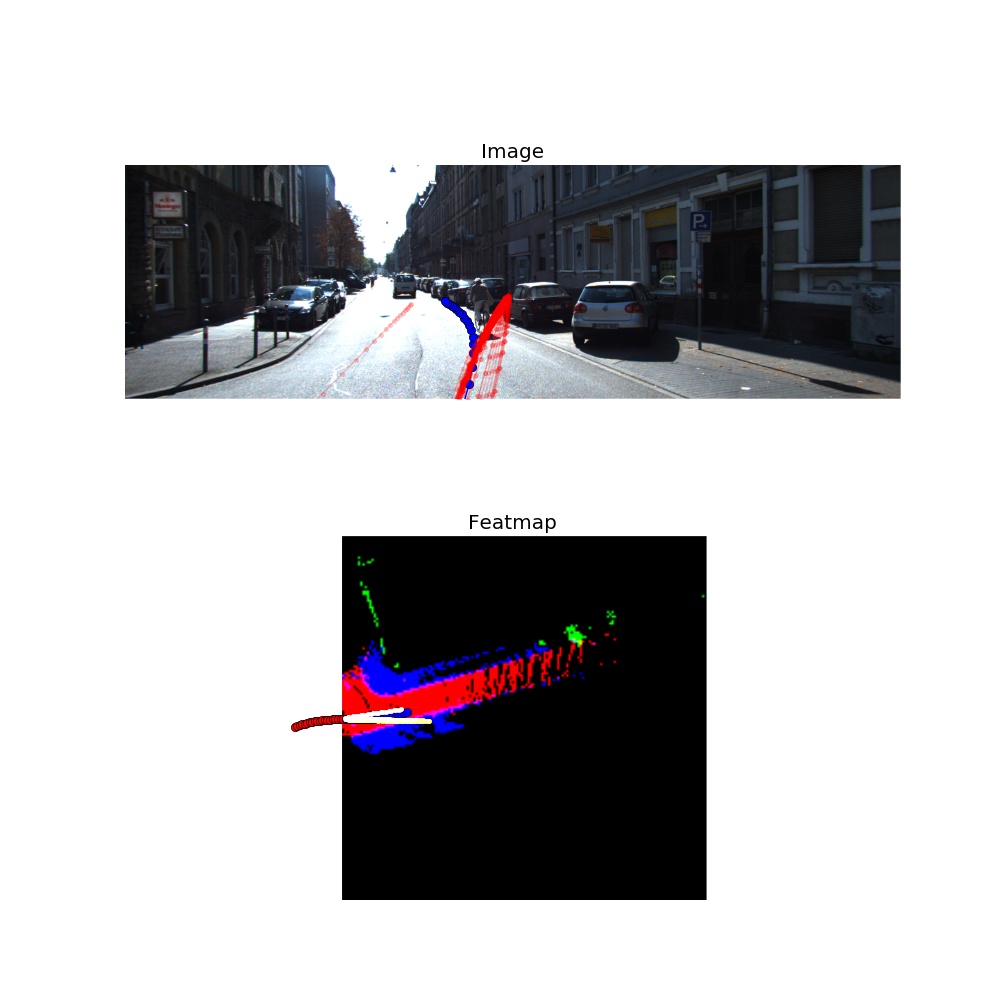}&
\includegraphics[width=0.33\linewidth,trim=80mm 210mm 100mm 58mm,clip]{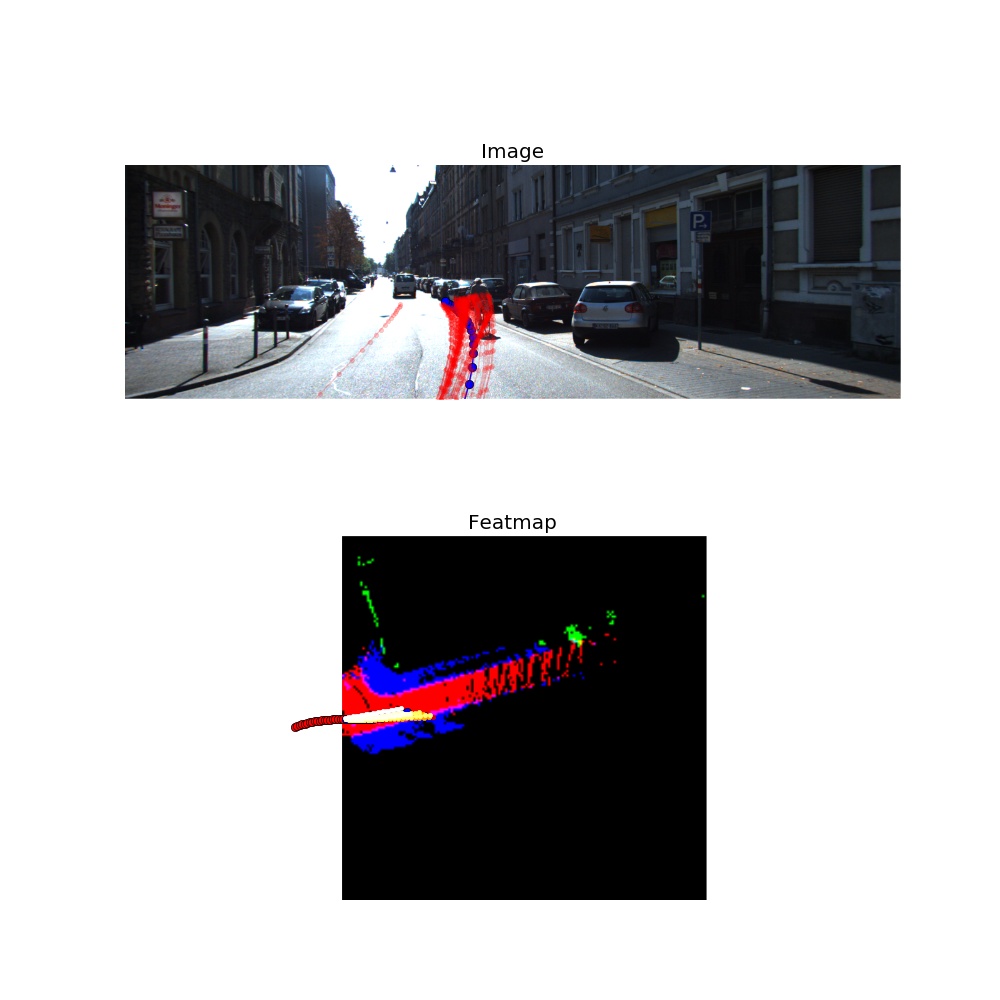}&
\includegraphics[width=0.33\linewidth,trim=80mm 210mm 100mm 58mm,clip]{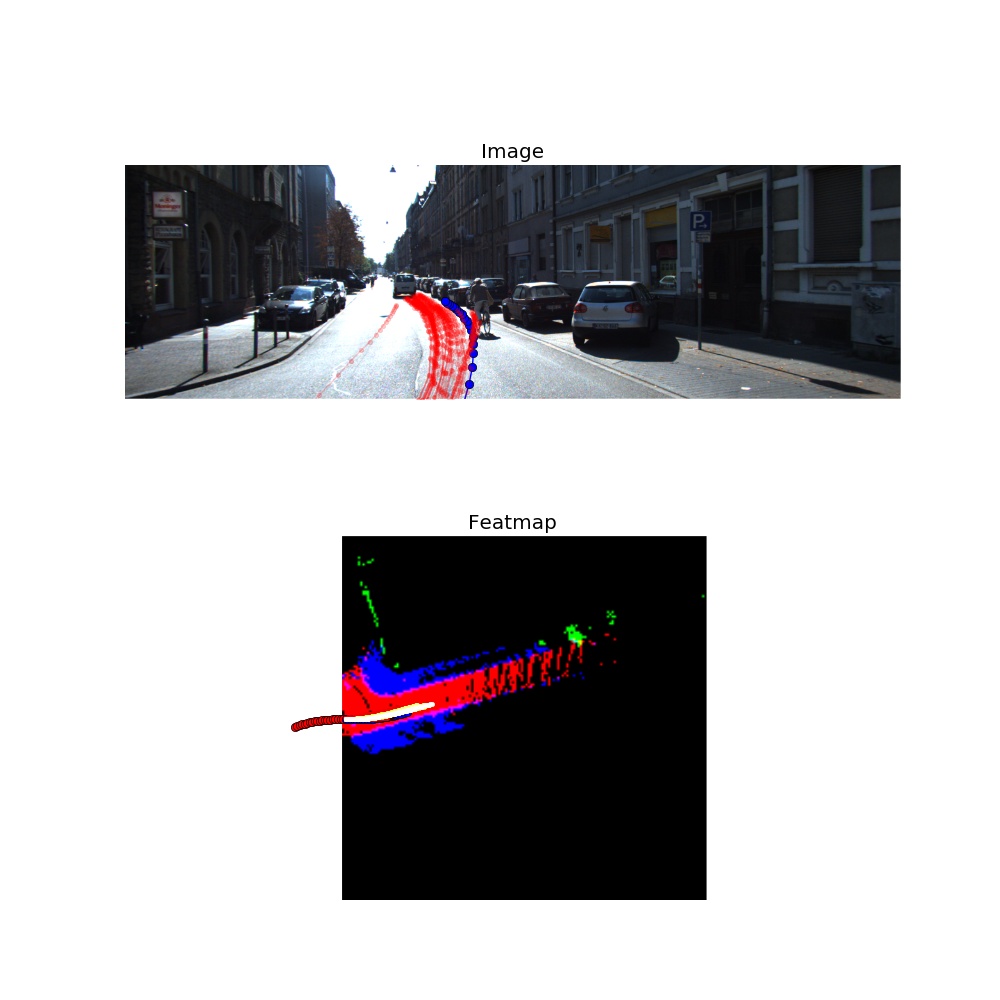}\\
\small Iteration 0 & \small Iteration 1 & \small Iteration 3 \\
\end{tabular}
\caption{Improved \textit{DESIRE-SI} prediction samples (\textcolor{red}{red}) over iterations. Iterative regression refines the predictions closer to the ground truth future trajectory (\textcolor{blue}{blue}) matching with scene context.}
\label{fig:iter}
\vspace{-4mm}
\end{figure}

Table~\ref{tab:quant} and Fig.~\ref{fig:quant} compare the oracle prediction errors\footnote{The maximum error in Table~\ref{tab:quant} might be different from Fig.~\ref{fig:quant} due to the test examples without ground truth labels at 4 seconds in the future.} of various methods. We present L2 distance error for both datasets and miss-rate with 1$m$ threshold for KITTI only, as trajectories in SDD are defined in image pixel space. Note that \textit{Linear}, \textit{RNN ED}, and \textit{RNN ED-SI} output a single prediction, thus their results are shown as horizontal lines. \textit{CVAE} samples are sorted randomly without confidence values. 

\begin{figure}
\vspace{-8mm}
\centering
\scriptsize
\includegraphics[width=1\linewidth,trim=21mm 113mm 20mm 5mm,clip]{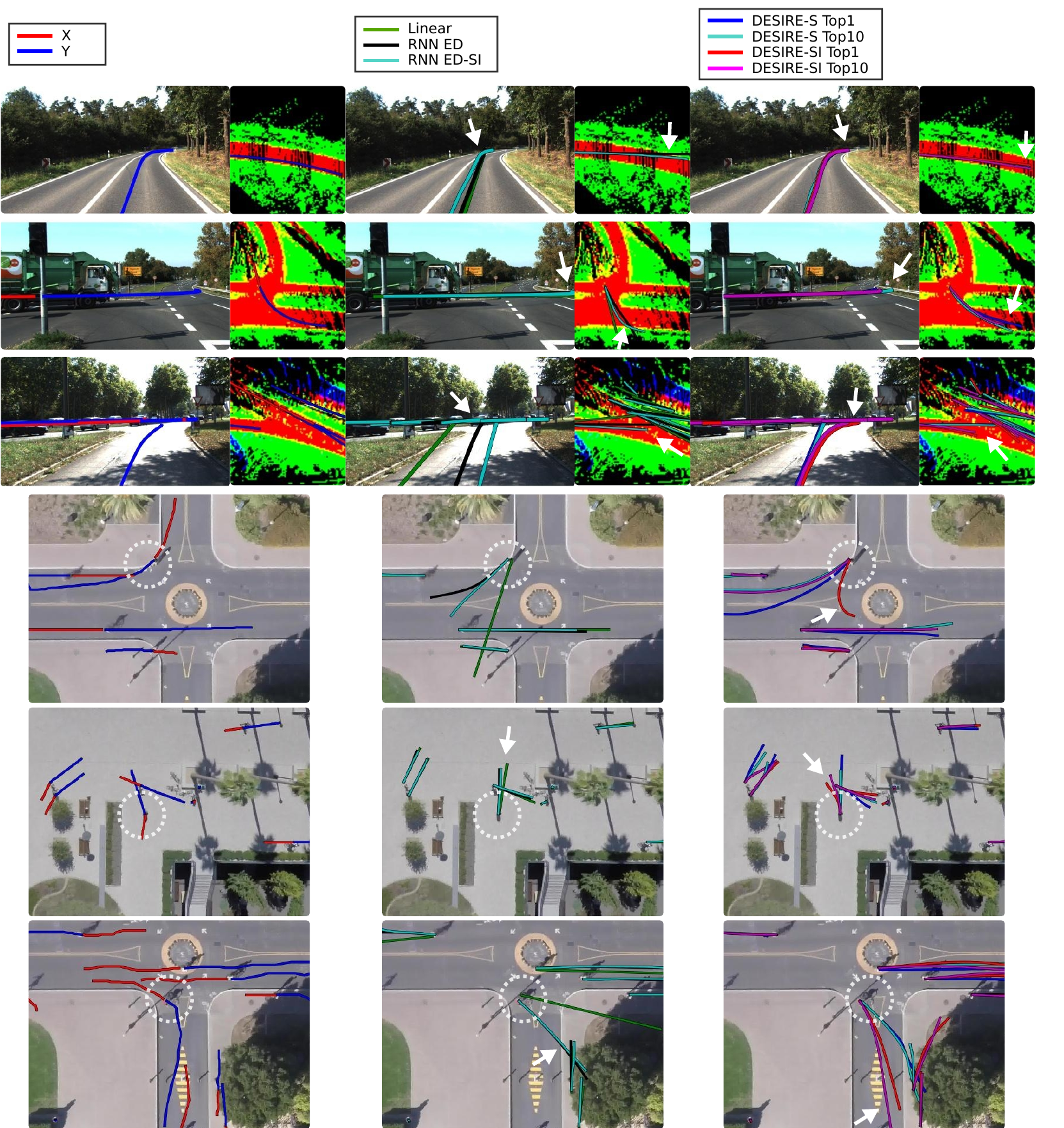}\\
\small (a) GT \quad\quad\quad\enspace\enspace\enspace\small (b) Baselines \enspace\quad\quad\quad\small (c) DESIRE
\caption{\textbf{KITTI results} (top 3 rows): The row 1\&2 in (b) show highly \textit{reactive} nature of \textit{RNN ED-SI} (\ie, prediction turns after it hits near non-drivable area). On the contrary, \textit{DESIRE} shows its long-term prediction capability by considering potential future rewards. \textit{DESIRE-SI} also produces more convincing predictions in the presence of other vehicles. \textbf{SDD results} (bottom 3 rows): The row 4 shows the multi-modal nature of the prediction problem. While the cyclist is making a right turn, it is also possible that he turns around the round-about (denoted with arrow). \textit{DESIRE-SI} predicts such equally possible future as the top prediction, while covering the ground truth future within top 10 predictions. The row 5\&6 also show that \textit{DESIRE-SI} provides superior predictions by reasoning about both static and dynamic scene contexts.}
\label{fig:kitti_qual}
\vspace{-6mm}
\end{figure}

\noindent\textbf{Baselines}: \textit{RNN ED} performs significantly better than \textit{Linear} since it can learn non-linear motion. We observe that \textit{RNN ED-SI} performs worse than \textit{RNN ED} on the KITTI since the model learns to behave \emph{reactive} (see Fig.~\ref{fig:kitti_qual}). This might be due to the small size of the dataset, which makes it hard to learn predictive CNN/interaction features (\ie, features need to have high capacity to encode long-term information). On the contrary, \textit{RNN ED-SI} significantly outperforms \textit{RNN ED} on SDD dataset since SDD is much bigger and has a large number of interactions among agents. 

\noindent\textbf{Proposed models}: With a single random sample (\textit{CVAE} 1 in Table~\ref{tab:quant}), \textit{CVAE} performs worse than \textit{RNN ED} since \textit{RNN ED} directly optimizes for L2 distance during training. Given more than few samples (e.g., \textit{CVAE} $10\%$ in Table~\ref{tab:quant}), \textit{CVAE} outperforms \textit{RNN ED} quickly on both datasets, which confirms the multi-modal nature of the prediction problem. \textit{DESIRE-X-IT0} without iterative regression properly ranks the random \textit{CVAE} samples achieving lower error with few samples. Note that \textit{DESIRE-X-IT0} only ranks the samples without regression, thus achieves the same error as used all samples, \ie, at Top K ratio of $1.0$ in Fig.~\ref{fig:quant}. As we iterate over, the outputs get refined and achieve smaller oracle error (\ie, \textit{DESIRE-X10\%-IT0} vs. \textit{DESIRE-X10\%-IT4}). Fig.~\ref{fig:iter} shows an example of the iterative feedback. Finally, we observe that considering the interaction between agents further helps to achieve lower error. The difference between \textit{DESIRE-S} and \textit{DESIRE-SI} is smaller in KITTI experiment, since KITTI has only few interactions between cars. However, we observe clear improvement on the SDD dataset since there are rich set of scenes with interactions between agents. Although our model with top 1 sample (\textit{DESIRE Best}) achieves higher error compared to the direct regression baselines, using a few more samples yields much better prediction accuracy (\ie, \textit{DESIRE} $10\%$). Note that direct regression models with lower error are not necessarily better if averaging various futures (\eg, going straight). We believe that in some applications, probabilistic prediction over a variety of outcomes is more desirable than a single MAP prediction.
For both datasets, \textit{DESIRE} achieves error on par with best baselines using as little as top $2$ samples of \textit{DESIRE-SI-IT4} predictions (see Fig.~\ref{fig:quant}). Qualitative results are presented in Fig.~\ref{fig:kitti_qual} and in the supplementary material.

\begin{table}
\vspace{-8mm}
    \scriptsize
    \centering
    \begin{tabular}{llllll}
        \toprule
        Method & 1.0 (sec) & 2.0 (sec) & 3.0 (sec) & 4.0 (sec)  \\
        \hline
        \multicolumn{5}{c}{KITTI (error in meters / miss-rate with 1 $m$ threshold)}\\
        \hline
        \textit{Linear} & 0.89 / 0.31 & 2.07 / 0.49 & 3.67 / 0.59 & 5.62 / 0.64 \\ 
        \textit{RNN ED} & 0.45 / 0.13 &  1.21 / 0.39 &  2.35 / 0.54 &  3.86 / 0.62 \\ 
        \textit{RNN ED-SI} &  0.56 / 0.16 &  1.40 / 0.44 &  2.65 / 0.58 &  4.29 / 0.65 \\
        \hline
        \textit{CVAE 1} & 0.61 / 0.22 &  1.81 / 0.50 &  3.68 / 0.60 &  6.16 / 0.65 \\ 
        \textit{CVAE 10\%} & 0.35 / 0.06 &  0.93 / 0.30 &  1.81 / 0.49 &  3.07 / 0.59  \\
        \hline
        \textit{DESIRE-S-IT0 Best} & 0.53 / 0.17 &  1.52 / 0.45 &  3.02 / 0.58 &  4.98 / 0.64\\
        \textit{DESIRE-S-IT0 10\%} & 0.32 / 0.05 &  0.84 / 0.26 & 1.67 / 0.43 & 2.82 / 0.54\\
        \textit{DESIRE-S-IT4 Best} & 0.51 / 0.15 &  1.46 / 0.42 &  2.89 / 0.56 & 4.71 / 0.63\\
        \textit{DESIRE-S-IT4 10\%} & \textbf{0.27} / \textbf{0.04} &  \textbf{0.64} / 0.18 &  \textbf{1.21} / 0.30 &  2.07 / 0.42\\
        \textit{DESIRE-SI-IT0 Best} & 0.52 / 0.16 &  1.50 / 0.44 & 2.95 / 0.57 &  4.80 / 0.63\\
        \textit{DESIRE-SI-IT0 10\%} & 0.33 / 0.06 &  0.86 / 0.25 &  1.66 / 0.42 &  2.72 / 0.53\\
        \textit{DESIRE-SI-IT4 Best} & 0.51 / 0.15 &  1.44 / 0.42 &  2.76 / 0.54 &  4.45 / 0.62\\ 
        \textit{DESIRE-SI-IT4 10\%} & 0.28 / 0.04 &  0.67 / \textbf{0.17} &  1.22 / \textbf{0.29} &  \textbf{2.06} / \textbf{0.41} \\
        \hline
        \multicolumn{5}{c}{SDD (pixel error at $1/5$ resolution)} \\
        \hline
        \textit{Linear} &  2.58 &  5.37 &  8.74 &  12.54 \\ 
        \textit{RNN ED} & 1.53 &  3.74 &  6.47 &  9.54 \\ 
        \textit{RNN ED-SI} &  1.51 &  3.56 &  6.04 &  8.80 \\
        \hline
        \textit{CVAE 1} & 2.51 &  6.01 &  10.28 &  14.82\\
        \textit{CVAE 10\%} & 1.84 &  3.93 &  6.47 &  9.65 \\ 
        \hline
        \textit{DESIRE-S-IT0 Best} & 2.02 &  4.47 &  7.25 &  10.29\\
        \textit{DESIRE-S-IT0 10\%} & 1.59 &  3.31 &  5.27 &  7.75 \\
        \textit{DESIRE-S-IT4 Best} & 2.11 &  4.69 &  7.58 &  10.66 \\
        \textit{DESIRE-S-IT4 10\%} & 1.30 &  2.41 &  3.67 &  5.62  \\
        \textit{DESIRE-SI-IT0 Best} & 2.00 &  4.41 &  7.18 &  10.23 \\
        \textit{DESIRE-SI-IT0 10\%} & 1.55 &  3.24 &  5.18 &  7.61 \\
        \textit{DESIRE-SI-IT4 Best} & 2.12 &  4.69 &  7.55 &  10.65 \\ 
        \textit{DESIRE-SI-IT4 10\%} & \textbf{1.29} &  \textbf{2.35} &  \textbf{3.47} &  \textbf{5.33} \\
        \bottomrule
    \end{tabular}
    \caption{Prediction errors over future time steps on KITTI and SDD datasets. Our method, \textit{DESIRE-IT4}, achieves by far the lowest top $10\%$ error, addressing the multimodal nature of the task effectively.}
\label{tab:quant}
\vspace{0mm}
\end{table}

\begin{table}
    \vspace{-4mm}
    \scriptsize
    \centering
    \begin{tabular}{lllll}
        \toprule
        Method & \multicolumn{4}{c}{K (the number of prediction samples)} \\
        & 25 & 50 & 100 & 200 \\
        \hline
        \textit{DESIRE-S-IT4 Best} & 4.87 & 4.71 & 4.81 & 4.70 \\
        \textit{DESIRE-S-IT4 top$20$} & 2.03 & 2.04 & 1.99 & 1.96 \\
        \bottomrule
    \end{tabular}
    \caption{Prediction errors of \textit{DESIRE-S-IT4} on KITTI at $4s$ for varying K. The best sample errors remain similar, while top 20 oracle errors decrease slightly as K increases.}
\label{tab:ablative_K}
\vspace{0mm}
\end{table}

\begin{table}
\vspace{-4mm}
    \scriptsize
    \centering
    \begin{tabular}{llll}
        \toprule
        Method & \multicolumn{3}{c}{Time length for past (sec)} \\
        & 1.0 & 2.0 & 4.0 \\
        \hline
        \textit{DESIRE-S-IT4 Best} & 4.94 & 4.71 & 4.78 \\
        \textit{DESIRE-S-IT4 10\%} & 2.11 & 2.07 & 2.05 \\
        \bottomrule
    \end{tabular}
    \caption{Prediction errors of \textit{DESIRE-S-IT4} on KITTI at $4s$ for varying time length for past trajectory. The model trained with $1s$ past slightly worse than ours ($2s$), showing that 2 second past contains enough cues to encode motion context. Note also that prior works adopt similar past lengths (2.8s in~\cite{alahisocial, robicquet2016learning})}
\label{tab:ablative_iota}
\vspace{-4mm}
\end{table}

\noindent\textbf{Ablative study}: We conduct further experiments for varying K and past length to supplement the main experiments and report the results in Table~\ref{tab:ablative_K} and Table~\ref{tab:ablative_iota}.

%% file: conclusion.tex
\section{Conclusion}
We introduce a novel framework \textit{DESIRE} for distant future prediction of multiple agents in complex scene. The model incorporates both static and dynamic scene contexts with a deep IOC framework and produces stochastic, continuous, and time-profiled long-term predictions that can effectively account for the uncertainty in the future prediction task. Our empirical evaluations on driving and surveillance scenarios demonstrate clear improvement over other baselines. For future work, we believe that our model can be further improved on larger datasets and be applied to various robotics applications with a direct use of perspective images. 

\vspace{-2mm}
\section*{Acknowledgement}
This work was part of N. Lee's summer internship at NEC Labs America and also supported by the EPSRC, ERC grant ERC-2012-AdG 321162-HELIOS, EPSRC grant Seebibyte EP/M013774/1 and EPSRC/MURI grant EP/N019474/1.